\documentclass[runningheads]{llncs}

\usepackage{eccv}

\usepackage{eccvabbrv}
\usepackage{graphicx}
\usepackage{booktabs}

\usepackage[accsupp]{axessibility}  

\usepackage{wrapfig}
\usepackage{multirow}

\usepackage{hyperref}

\usepackage{orcidlink}

\begin{document}

\title{Pixels to Keys: Exploring Spatial and Motion Cues in Gameplay Inverse Dynamics}
\titlerunning{Pixels to Keys}

\author{
Abhishek Pillai\inst{1} \and
Ekta Prashnani\inst{1} \and
Joohwan Kim\inst{1} \and
Iuri Frosio\inst{1} 
}

\authorrunning{A.~Pillai et al.}

\institute{
NVIDIA\\
\email{\{abpillai,eprashnani,sckim,ifrosio\}@nvidia.com}
}

\maketitle
\begin{abstract}
  Video games offer scalable environments for studying perception and control in embodied agents.
  Abundant online gameplay videos could supply demonstrations, but they rarely include player inputs for training.
  Inverse Dynamics Models (IDMs) have thus been proposed to infer inputs from frames.
  Large (up to 1B parameters) IDMs trained on $\sim$1K-2K gameplay hours demonstrate feasibility and cross-environment generalization at this scale, but researchers do not clarify what the key components are to recover individual actions and often report only aggregate accuracy that can mask rare-action failures.
  We study the problem in a data-constrained scenario to evaluate how spatial motion features, model architectures, and training objectives affect an IDM's outcome and we analyse our models on per-key and balanced metrics such as $F_1^{macro}$.
  Our experiments on Trackmania highlight the importance of factors like the model architecture and motion flow extraction in preprocessing, while also showing the limits of evaluation through unbalanced metrics.
  The application of the same architecture and training recipe to Cyberpunk 2077 reveals uneven performance across game mechanics.
  Our per-action evaluation and failure analysis highlight ambiguities from camera motion, delayed effects and imbalanced key-press frequencies that call for explicit modeling of 3D scene structure, long-term state and the adoption of proper losses in future implementations.
\end{abstract}

\section{Introduction}
\label{sec:intro}

\begin{wrapfigure}{R}{0.53\textwidth}
{
\centering
\includegraphics[width=\linewidth]{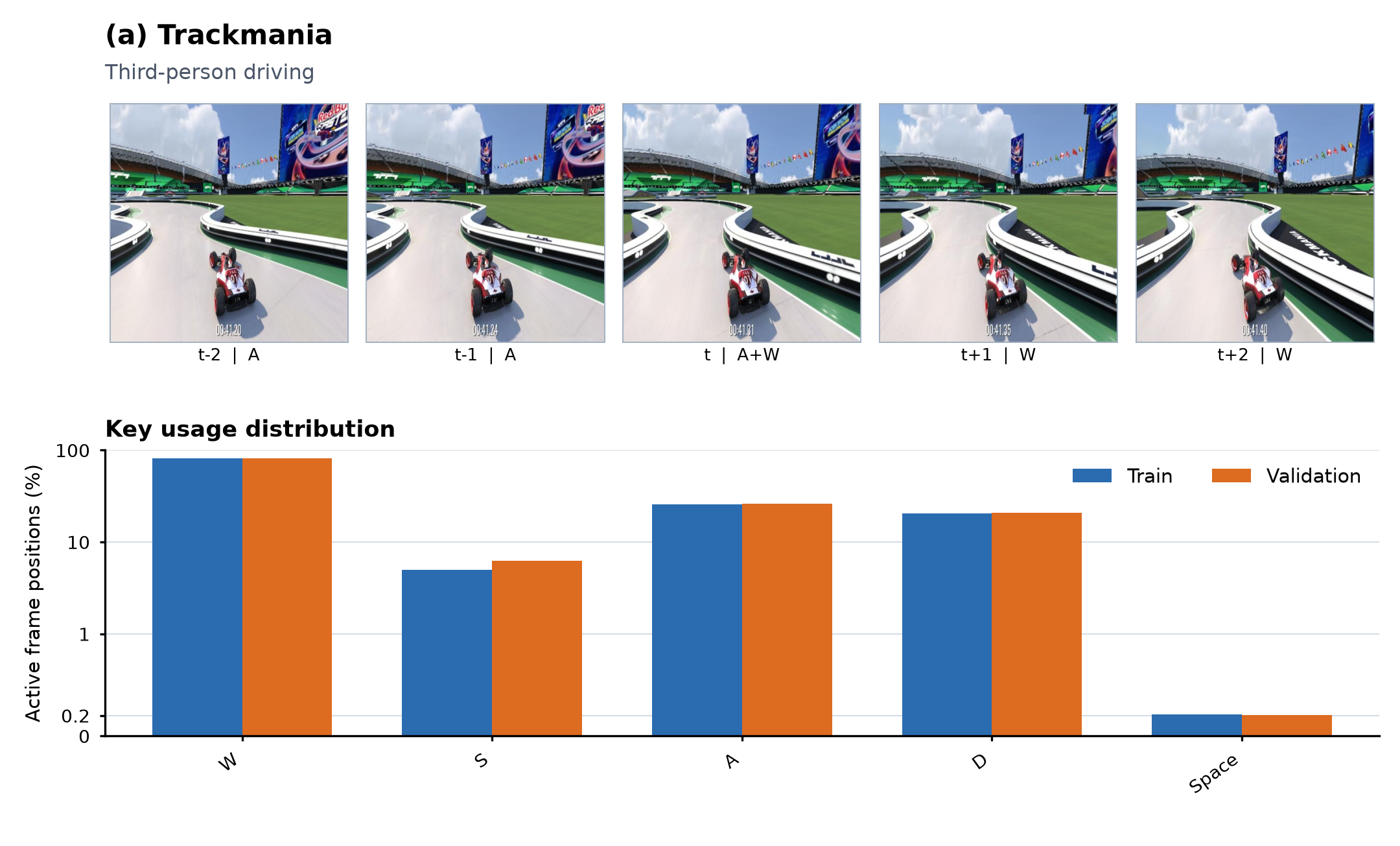} \\
\includegraphics[width=\linewidth]{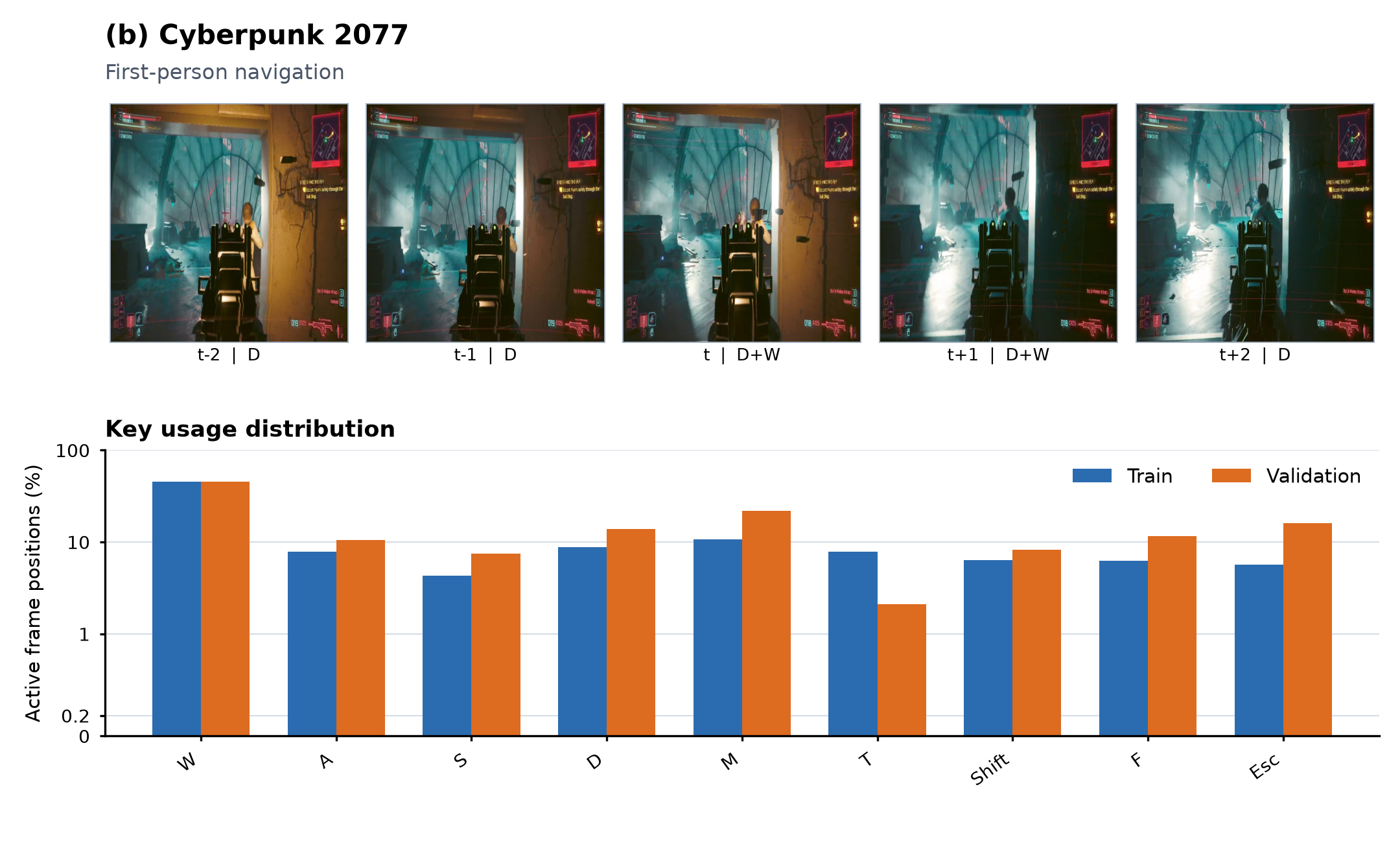}
\caption{Sample sequences and key-press distributions for Trackmania (a) and Cyberpunk 2077 (b). Distributions are on a log scale: key usage is highly imbalanced.}
\label{fig:game-datasets}
}
\end{wrapfigure}

Beyond their commercial importance, video games are used as testbeds for perception, sequential decision making and embodied control~\cite{berner2019dota}\cite{li2024okami}\cite{mnih2013atari}. Learning from gameplay requires pairs of frames and actions issued by human or digital players to train agents with methods like Behavioral Cloning (BC)~\cite{baker2022video}, Reinforcement Learning (RL)~\cite{Sutton1998}, or to gain understanding of visual-motor control.


Despite abundant gameplay videos available on the internet, only a few of these are annotated with \emph{Keyboard} and \emph{Mouse} (KBM) actions.
Manual annotation at scale is practically impossible, whereas the acquisition of new datasets with frames labeled with KBM data is costly and time consuming.
Therefore we study the possibility of automatically annotating such videos with KBM labels.
The problem is far from trivial for several reasons, including actions with delayed effects (e.g., inertia affecting camera motion without any steady input); also, different actions may produce the same visual results, whereas certain actions may produce poorly visible features (e.g., a car turning while driving against a wall).
Additionally, the action distribution may be highly imbalanced (see Fig.~\ref{fig:game-datasets}).
Disentangling camera motion from character motion is yet another complication. 

Existing works establish this idea at scale.
VPT~\cite{baker2022video} is a 0.5B parameter IDM trained on 1,962 hours of labeled Minecraft data and used to annotate 70K hours of web video.
D2E~\cite{choi2025d2e} uses a generalist model with 1B parameters trained on 1.3K hours of data (259 hours of human demonstration, 1K hours of pseudo-labeled gameplay).
These systems recover broad KBM action spaces and demonstrate the value of scale, but the question we intend to answer is deliberately smaller: \textit{before broad action recovery, how well can we recover navigation inputs from purely visual data?}
By answering this we highlight constraints that may become significant at a larger scale and in applications such as: annotation of online recordings, latency correction in poorly controlled acquisition setups, or the adoption of KBM annotation as an auxiliary task in complex training scenarios.

Our contributions are:
\begin{enumerate}
    \item For Trackmania, a $3^{rd}$ person view and mechanics video game, we isolate the effects of different factors (including frame resolution, use of pretrained embeddings, loss terms and prediction size) and highlight that model architecture, optical flow in input, and contrastive learning provide the most significant advantages.
    \item We highlight the importance of adopting per-key and unweighted metrics for proper evaluation of IDMs.
    \item We identify limitations, open problems, and guidelines for future research by applying the same model and training recipe to Cyberpunk 2077 ($1^{st}$ person view). We analyze, on both games, the spatial action-query attention map, to show, for instance, the importance of equipping the IDM with the capability of disentangling the camera motion from character motion.
\end{enumerate}

\section{Related Work}
\subsubsection{Large-Scale IDMs and Action-Level Evaluation:}
Latent-action IDM techniques remove the labeled corpus altogether: for instance, LAPO~\cite{schmidt2024lapo} trains a latent IDM and forward model jointly on 8M action-free Procgen frames; Genie~\cite{bruce2024genie} scales this approach to 30K\,hours of video with a 300M-parameter latent model.
Both recover a learned action space, later mapped using expert labels, rather than true actions: in other words, the reconstructed set of actions guarantees that the agent behaves like in the training examples, not that the set of key-presses is the same.
Here we tackle a complementary question: \emph{when the action space is fixed and discrete, how accurately can true key-presses be recovered?}
In this, we are closer to D2E~\cite{choi2025d2e}, a 1B-parameter generalist agent trained on 259 hours of human demonstration and 31 games plus 1K hours of pseudo-labeled gameplay, scored on 50\,ms non-overlapping event bins.
When zero-shot on unseen titles, its keyboard accuracy falls to 63\% (Battlefield 6) and 28\% (Ogu and the Secret Forest): scale alone does not solve the key-press reconstruction problem.

We also found that the quantitative evaluation of IDMs is often aggregated over keys: VPT reports a single 90.6\% key-press accuracy; D2E reports one keyboard accuracy per game.
Per-class breakdowns are rare, and mostly reserved to world-model controllability or GUI-agent evaluations~\cite{guo2025mineworld,zhang2025matrixgame,lu2025videoagenttrek}. When reported, the variance is severe: MineWorld's per-action F1 ranges from $\sim0.50$ (\texttt{drop}) to $\sim0.80$ (\texttt{forward/backward}).
In VideoAgentTrek, F1 for \texttt{press} is $0.14$, for \texttt{scroll} it reaches $0.86$, but aggregation over all actions has $F1=0.78$.
In other words, aggregation hides variance when the majority label dominates.
This aspect should not be neglected in training and evaluation: in Trackmania (Fig.~\ref{fig:game-datasets}a), W (\texttt{forward}) is $4\times$ more frequent than A/D (\texttt{left}/\texttt{right}), therefore an always-press-W predictor scores well on the aggregate while failing on rarer actions.
For this reason we study per-key metrics both for training and evaluation.

Researchers in robotics face a similar recovery problem for continuous actions: H2O~\cite{he2024h2o} and OKAMI~\cite{li2024okami} fit a parametric body model to human video, then retarget the estimated pose to humanoid joint angles.
Because that target is itself an estimate, it is validated by downstream task success rather than compared with ground truth; our labels are key-presses, so predictions are scored directly.

\subsubsection{Motion, Temporal and Semantic Space Design:}
Visual dynamics in games are complicated by 3D scene structure, elements that do not follow scene physics (such as HUD overlays) or real physics (such as magic items), and by the entanglement of the camera and character motion.
To provide better motion understanding to our model, we test augmentation of the RGB input with RAFT optical flow, which encodes estimated per-pixel displacement between adjacent frames.
Its all-pairs correlations and iterative refinement capture large camera and character motions, offering direct cues for key-press prediction~\cite{teed2020raft}.
We also study architectures that are specifically designed to handle videos and, as a consequence, motion.
In particular, we compare two model families: video transformers such as ViViT~\cite{arnab2021vivit} and TimeSformer~\cite{bertasius2021timesformer} that apply joint or factorized space-time attention to patch or tubelet tokens; and Hybrid models using a convolutional stem to extract local features and downsample $H\times W$ inputs before tokenization~\cite{xiao2021early,dai2021coatnet}.
To handle the complex semantics of video games, we also test whether pretrained DINOv3 embeddings~\cite{simeoni2025dinov3} (ViT-S/16) transfer to key-press estimation.
Lastly, BCE used by prior IDMs~\cite{baker2022video,choi2025d2e} can underperform on imbalanced and sparse multi-label targets, so we compare it with adaptive asymmetric loss~\cite{benbaruch2021asl}, soft-F1~\cite{benedict2022sigmoidf1}, and action-supervised contrastive initialization~\cite{khosla2020supcon} as alternative training objectives.

\section{Method}
\subsection{Problem Formulation}
\label{sec:problem-formulation}

We describe the case of the Trackmania video game with 5 navigation keys $\mathcal{K}=\{\mathrm{W},\mathrm{A},\mathrm{S},\mathrm{D},\mathrm{Sp}\}$, where $Sp$ stands for $Space$. The first four keys are used to accelerate / turn left / brake / turn right; the last one resets the car to the starting point (and it is therefore used only once per episode).
For Cyberpunk and other video games, the set of navigation keys may be different (Fig.~\ref{fig:game-datasets}), but generalization is trivial.
For direct comparison, we use the same five-key set $\mathcal{K}$ for Cyberpunk 2077.
Our model takes in input a sequence of $T_{in}$ RGB frames and we adopt $T_{in} = 5$ as a baseline, but generalization to other lengths or input modes (for instance with an additional optical flow channel), is again trivial.

Given a gameplay clip $\mathbf{X}=(\mathbf{x}_{t-2},\ldots,\mathbf{x}_{t+2})\in\mathbb{R}^{T_{in}\times3\times H\times W}$, our IDM predicts:
\begin{equation}
\hat{\mathbf{y}}_t=
[\hat{y}_t^{\mathrm{W}},\hat{y}_t^{\mathrm{A}},\hat{y}_t^{\mathrm{S}},\hat{y}_t^{\mathrm{D}},\hat{y}_t^{\mathrm{Sp}}]
\in[0,1]^5
\end{equation}
where $\hat{y}_t^{\mathrm{W}}$ is the estimated probability of pressing $W$ at time $t$, and multiple keys may be active simultaneously. At inference time, probabilities can be used to sample each key-press value or be thresholded at $0.5$ (our selection).

%

Since we found experimentally that neither a $2^5$-class joint-action formulation nor increasing clip length (from $T_{in}=5$ (0.25\,s  at 20\,Hz) to $T_{in}=16$ (0.80\,s at 20\,Hz)) improves performance, we retain independent outputs and $T_{in}=5$ in all our experiments.
Furthermore, some of our architectures (see section~\ref{sec:architectures}) predict not only the key-presses at time $t$, but the full key states for the complete input interval, $\{\hat{\mathbf{y}}_{\tau}\}_{\tau=t-2}^{t+2}$.
In the following, we refer to these as \emph{center-frame} (or $T_{out}=1$) versus \emph{all-frames} (or $T_{out}=5$) supervised models.

\subsection{Datasets}
\label{sec:datasets}
We use 2 gameplay datasets: the first one comprises Trackmania recordings, a $3^{rd}$ person driving video game; the second one comprises Cyberpunk 2077, a $1^{st}$ person navigation video game (see Fig.~\ref{fig:game-datasets}).
Frames were captured at 20Hz and $512\times512$ resolution. The size and splits of the two datasets are shown in Table~\ref{tab:datasets}.
Models are trained and evaluated per game without mixing datasets, using only sequences with active gameplay. Each split is sampled from different recordings.

We augment input frames using a random affine transform shared across the full sequence with probability $0.75$, scale in $[0.95,1.05]$ range, rotation in $[-7^\circ,7^\circ]$ range, and translation in $[-32,32]$ pixels range.
To address the key-press class imbalance (stats in Fig.~\ref{fig:game-datasets}) we sample sequences with replacement using inverse-frequency weights.
Let $a_{ik}=1$ if key $k$ is pressed at any time in sequence $i$ and $0$ otherwise, and $f_k$ the $k$'s frequency over the training set. 
In training, we sample sequence $i$ with frequency $w_i\propto1+\sum_k a_{ik}/(f_k+0.01)$; validation and testing are on their complete splits without augmentation or resampling. 

\begin{table}[tb]
\centering
\caption{Number of overlapping 5-frame sequences per dataset. Each split's sequences belong to different recordings that are disjoint.}
\label{tab:datasets}
\small
\setlength{\tabcolsep}{4pt}
\begin{tabular}{c|ccc}
Game & Training sequences & Validation sequences & Testing sequences \\
\hline
Trackmania & $30{,}000$ ($\sim25$\,mins) & $20{,}294$ ($\sim17$\,mins) & $27{,}984$ ($\sim23$\,mins) \\
Cyberpunk 2077 & $72{,}192$ ($\sim60$\,mins) & $25{,}600$ ($\sim21$\,mins) & $24{,}576$ ($\sim20$\,mins) \\
\end{tabular}
\end{table}

\subsection{Training}
\label{training_details}

\subsubsection{Optimization}
All experiments run on 1 NVIDIA RTX PRO 6000 GPU and use BF16 autocasting, with losses computed in FP32. We use AdamW~\cite{loshchilov2019decoupled, kingma2017adam} with weight decay $0.1$ and learning rate $5\times10^{-6}$; 5 linear warm-up epochs are followed by CosineAnnealingLR~\cite{loshchilov2017sgdr} to $10^{-7}$. Batch size is 128, dropout is $0.1$, and each RGB or optical-flow input element is independently masked with probability $0.20$. Ground-truth key states are always masked and are never provided to the model as input.
Training is performed for a number of epochs between 45 and 100, with validation computed at each epoch.
The model checkpoint with the highest validation macro-$F_1$ is saved for inference.

\subsubsection{Training Loss}
Given the strong imbalance in the key-press probabilities, we adopt a Soft-F1 loss for training, defined as:
\begin{eqnarray}
F^{\mathrm{soft}}_{1,k}=\frac{2TP_k}{2TP_k+FP_k+FN_k+10^{-3}}\\
\mathcal{L}_{\mathrm{F1}}=1-\frac{1}{5}\sum_{k=1}^{5}F^{\mathrm{soft}}_{1,k}
\end{eqnarray}
where $TP_k$, $FP_k$, and $FN_k$ are the soft True Positive, False Positive, and False Negative counts for key $k$, respectively, summed over each minibatch.


\subsection{Ablation study (primary factors)}
\label{sec:ablation_main}
In our experiments we have found \textbf{three} factors emerging as the main ones affecting the model's capability to label each frame with the correct set of pressed keys.
These are (i) the model architecture, (ii) the additional optical flow channels in input, and (iii) the contrastive training objective: these are described here.
Experiments on other factors with minor effects on the final result (i.e., frame resolution, BCE and asymmetric loss, pretrained embeddings) are detailed in  Appendix \ref{sec:ablation_secondary}.

\subsubsection{Architectures}
\label{sec:architectures}
We considered architectures whose input/output is defined in \ref{sec:problem-formulation}, but have different internal skeletons: Convolutional Neural Networks (CNN), Vision Transformers (ViT), and Hybrid Transformers (HT).

\paragraph{CNN:} 
A set of $T_{in}$ RGB frames is stacked into $3 \times T_{in}$ channels, followed by 3 convolutional layers ($3 \times T_{in} \xrightarrow{} 32$ (kernel/stride $4/4$), $32 \xrightarrow{} 16$ ($3/1$), and $16 \xrightarrow{} 4$ ($5/5$)) with LeakyReLU activations.
Adaptive average pooling then yields a $4\times6\times6=144$-dimensional vector, followed by MLP layers ($144 \xrightarrow{} 16 \xrightarrow{} 5 \times T_{in}$).
This model (CNN-512) produces an \emph{all-frames} output: it predicts key-presses over $T_{out}=5$ frames ($25$ logits) and has $16{,}685$ parameters.
The training loss is computed on the entire output sequence, from $t-2$ to $t+2$.

\paragraph{Spatiotemporal ViTs:}
Each RGB frame is split into $16\times16$ patches, yielding $32\times32=1{,}024$ tokens per frame and $1{,}024 \times T_{in}$ joint space-time tokens ($5{,}120$ total for $T_{in}=5$).
Spatial and frame embeddings are learned end-to-end and precede 2 transformer blocks with embedding dimension $384$, $4$ heads, and MLP width $1{,}536$; framewise mean pooling and a shared head then produce 5 logits per frame.
Like the CNNs, this model (ViT-E2E-512) produces \emph{all-frames} outputs ($T_{out}=5$) and training loss is computed over the full sequence ($t-2$ to $t+2$).

\paragraph{Hybrid Transformer (HT):}
Our HT model retains CNN-512's channel-stacked convolutional input: RGB contributes $3 \times T_{in}$ input channels. 
For $T_{in}=5$, HT-RGB has $15$ channels. 
A 4-stage convolutional stem (kernel $16$, stride $2$, padding $7$; GroupNorm and LeakyReLU) maps $C_{\mathrm{in}}\in\{15,23\}$ through channels $64\rightarrow128\rightarrow256\rightarrow256$ while reducing spatial size as $512\rightarrow256\rightarrow128\rightarrow64\rightarrow32$.
The $256\times32\times32$ output is flattened into $1{,}024$ tokens of dimension $256$ and processed by 4 transformer blocks (LayerNorm before attention and MLP, $8$ heads, MLP hidden dim. $1{,}024$).
5 learned queries of dimension $256$ cross-attend to the tokens, and a common $256\rightarrow1{,}024\rightarrow5$ head produces 5 logits per query.
This model (HT-RGB-E2E) has $\sim31.5$M parameters.
Unlike CNNs and ViTs, it is supervised over the keys pressed at time $t$ only (\emph{center-frame} architecture with $T_{out}=1$), on which the loss is computed.

\subsubsection{Additional motion flow in input} Optical flow can be provided as an additional input channel.
The rationale is that clean motion information in input may be used by the model to reconstruct the character and camera motion and thus facilitate the estimate of the key-presses in the input sequence.
We use a RAFT-Large~\cite{wang2024searaft} model to compute the motion (dx, dy) for each of the 4 adjacent frame pairs (for $T_{in}=5$), and concatenate the resulting 8 channels with the RGB data.
Our HT-RGBF-E2E model shares the same architecture as the HT-RGB model, but has $23$ input channels ($4$ flow fields, $8$ channels) instead of $15$.
It also has a comparable ($\sim31.5$M parameters) size.

\subsubsection{Contrastive embeddings} We use a contrastive training objective as described below. An additional head with a $256\rightarrow256\rightarrow128$ GELU projector is added to the HT-RGBF model to output $\ell_2$-normalized embeddings (notice that this is used in training only, thus the size of the model at inference time won't change).
Each training sequence is then augmented in 2 different ways, using (beyond the default affine transform noted in section \ref{sec:datasets}) random contrast change in $[0.85,1.15]$ range, random brightness change in $[-0.15,0.15]$ range, and Gaussian noise with zero mean and $\sigma=0.0375$.
For a batch of $B$ sequences, the 2 augmentations yield $2B$ embeddings $z_i\in\mathbb{R}^{128}$ with $\|z_i\|_2=1$.
Let $y_i\in\{0,1\}^5$ be the center-action label of sequence $i$ and $A(i)=\{1,\ldots,2B\}\setminus\{i\}$ contain all non-self comparisons and $P(i)=\{p\in A(i):y_p=y_i\}$ contain its positive action matches.
The contrastive loss is then defined as:
\begin{equation}
\mathcal{L}_{\mathrm{SupCon}}=-\frac{1}{2B}\sum_{i=1}^{2B}\frac{1}{|P(i)|}
\sum_{p\in P(i)}\log\frac{\exp(z_i^\top z_p/\tau)}
{\sum_{a\in A(i)}\exp(z_i^\top z_a/\tau)},\qquad \tau=0.20.
\end{equation}
As embeddings are normalized, $z_i^\top z_j$ is cosine similarity; $\tau$ is softmax temperature. The loss averages over all $2B$ embeddings and their $|P(i)|$ positives, while the paired augmentation guarantees $|P(i)|\geq1$. All $N(i)=A(i)\setminus P(i)$ embeddings with $y_a\neq y_i$ are negatives: their presence in the denominator penalizes high negative similarity, separating them from positives.
When training with contrastive loss, we first run 20 epochs to minimize $\mathcal{L}_{\mathrm{SupCon}}$ and learn a continuous embedding space.
After that, the prediction head is reinitialized and training continues for another 25 epochs with the prescribed loss only. This produces the HT-RGBF-PT-F1 model reported in Table~\ref{tab:results-ablations-main-copy}.

\section{Experiments}
\subsection{Evaluation metrics}%
Let $y_t^k$ and $\hat y_t^k$ be the ground-truth and predicted press probability of key $k$ at timestep $t$. In validation, we threshold each $\hat y_t^k$ at 0.5.
For each key, the true/false positives/negatives $TP_k$, $FP_k$, $TN_k$, and $FN_k$ count positions with $(\hat y_t^k,y_t^k)=(1,1)$, $(1,0)$, $(0,0)$, and $(0,1)$, respectively.
The same terms (without $k$ index) indicate the overall true/false positives/negatives, computed for all the keys over the entire validation or test dataset.
To evaluate and compare the different models trained in our ablation study, we utilize the following metrics:

\begin{enumerate}
\setlength{\itemsep}{1pt}
\item $\mathrm{\text{Accuracy (Acc.)} = \frac{TP+TN}{TP+TN+FP+FN}}$ 

\item $F_{1,k}=\frac{2TP_k}{2TP_k+FP_k+FN_k}$

\item $F_1^{\mathrm{macro}}=\frac{1}{5}\sum_{k=1}^{5}F_{1,k}$ 

\item $F_1^{\mathrm{micro}}=\frac{2\sum_k TP_k}{2\sum_k TP_k+\sum_k FP_k+\sum_k FN_k}$ 
\end{enumerate}

Accuracy is widely reported in previous works, but the majority class and inactive labels can dominate $\text{TP+TN}$, severely inflating performance.
The second one, $F_{1,k}$, is the $F_1$ score (that balances precision and recall) computed per key: it is therefore less biased in case of an imbalanced dataset, where the key $k$ is only rarely or very often pressed.
Since all $k$ values have to be inspected, $F_{1,k}$ is on the other hand tedious to review and hardly usable for ranking. $F_1^{\mathrm{macro}}$ is the mean over all $F_{1,k}$: it weighs down the imbalanced class by ignoring the key frequencies.
It helps reveal rare-key performance and it is our primary metric.
Notice in fact that, when evaluating video games, rare key-presses
often play a fundamental role in the economy of the game: e.g., shooting may be less frequent than moving forward, but contrarily to navigation, precise and accurate shooting is required to stay alive.
$F_1^{\mathrm{micro}}$ generalizes the traditional $F_1$ score used in binary problems to the multilabel case: it pools counts across all keys, summarizing dataset-level recovery; it may still be affected by key frequency imbalance.





\subsection{Results}
\label{tab:configs-42}

Table~\ref{tab:results-ablations-main-copy} shows metrics on the test datasets for the main experiments; within each epoch budget, checkpoints are selected by $F_1^{macro}$ on the validation split.
Configuration columns report the architecture, $T_{out}$, resolution, flow input, initialization strategy, loss used for training, and number of training epochs.
Training and validation curves are in Appendix~\ref{app:training-validation-curves}.
All results are reported on Trackmania, except the final row, where the HT-RGBF-PT-F1 model and training recipe is applied to Cyberpunk 2077.
Complete configurations for the main experiments and secondary ablations are reported in Table~\ref{tab:architectures} in the Appendix.



\begin{table}[tb]
\centering
\caption{Accuracy Acc. and $F_1$ metrics on the test datasets;
$20+25$ denotes 20 contrastive and 25 supervised epochs.
All rows refer to Trackmania except the last one marked $^*$, which uses Cyberpunk 2077. \textbf{Bold} and \underline{underlined} mark the highest and second-highest distinct Trackmania values in each metric column, respectively. For Cyberpunk 2077 we report for comparison only navigation keys that are in common with Trackmania, with Accuracy, $F_1^{\mathrm{macro}}$, and $F_1^{\mathrm{micro}}$ over that set.}
\label{tab:results-ablations-main-copy}
\label{tab:results-ablations}
\resizebox{\textwidth}{!}{%
\begin{tabular}{cccccccc|cccccccc}
\multicolumn{1}{c}{\multirow{2}{*}{Model}} & \multicolumn{7}{c|}{Configuration} & \multicolumn{1}{c}{\multirow{2}{*}{Acc.}} & \multicolumn{7}{c}{$F_1$} \\
& $T_{\mathrm{out}}$ & Arch. & Res. & Flow & Initialization & Loss & Epochs & & micro & macro & W & A & S & D & Sp \\
\hline
CNN-512      & $5$ & CNN & $512^2$ & N & none & soft-F1 & 100 & .489 & .499 & .367 & .902 & .434 & .075 & .415 & .008 \\
\hline
ViT-E2E-512  & $5$ & ViT & $512^2$ & N & none & soft-F1 & 100 & .887 & .817 & .637 & .907 & .698 & .567 & .755 & .256 \\
\hline
HT-RGB-E2E   & $1$ & HT  & $512^2$ & N & none & soft-F1 & 50 & \underline{.929} & \underline{.880} & .746 & \textbf{.921} & \underline{.837} & .648 & \underline{.823} & .502 \\
HT-RGBF-E2E  & $1$ & HT  & $512^2$ & Y & none & soft-F1 & 50 & \underline{.929} & .878 & \underline{.791} & \underline{.920} & .829 & \underline{.655} & .822 & \textbf{.728} \\
\hline
HT-RGBF-PT-F1    & $1$ & HT & $512^2$ & Y & contrastive & soft-F1 & $20+25$ & \textbf{.938} & \textbf{.895} & \textbf{.793} & \underline{.920} & \textbf{.869} & \textbf{.688} & \textbf{.870} & \underline{.618} \\
\hline
CP-HT-RGBF-PT-F1* & $1$ & HT & $512^2$ & Y & contrastive & soft-F1 & $20+25$ & .911 & .715 & .617 & .786 & .571 & .710 & .543 & .473 \\
\end{tabular}%
}
\end{table}

\subsection{Quantitative Analysis}

Different architectures are compared in the first three rows of Table~\ref{tab:results-ablations}.
While an \emph{all-frames} CNN model's $F_1^{\mathrm{macro}}$ is as low as $0.367$, \emph{all-frames} ViTs achieve $0.637$ using the same frame input resolution, training loss function and number of epochs.
The \emph{center-frame} HT architecture in the third row beats both of them with $F_1^{\mathrm{macro}} = 0.746$.
Despite the experiment being performed on one seed only, the high difference in $F_1^{\mathrm{macro}}$ together with the fact that all other metrics are also higher for HT (when compared to CNN and ViT) indicate with strong evidence the superiority of the HT architecture over CNN and ViT.

The comparison of the third (HT-RGB-E2E) and fourth (HT-RGBF-E2E) rows in the same Table allows estimating the effect of the additional motion flow in input.
We notice that $F_1^{\mathrm{macro}}$ improves by approximately $6\%$ relative (from $0.746$ to $0.791$) when motion information is provided.
Other metrics remain more or less in the same ballpark, with the exception of $F_{1,Sp}$ that shows a $45\%$ increase.
Also in this case the improvement is thus numerically relevant, although not as large as the one registered for the case of different architecture: more experiments with different seeds may be needed to establish more precisely the advantage of the additional motion flow in input.

The comparison between the HT-RGBF-E2E and HT-RGBF-PT-F1 isolates the effect of contrastive initialization.
The HT-RGBF-PT-F1 model shows a consistent improvement in all core navigation metrics with the only exception of $F_{1,Sp}$.
Even if the improvement is on average not large, the fact that it is common to the majority of the metrics suggests that the improvement over HT-RGBF-E2E is real.
This highlights, likewise in other contexts, the importance of learning semantically significant and smooth representations of the input patches in video games.
This claim becomes even more significant when considering that embedding extraction performed with models pretrained on real-world data (see the DINOv3 case in the Appendix) did not provide any improvement in our experiments.
In other words, the best performance is achieved when the embedding space is learned end-to-end (and thus visually and semantically meaningful in the video-game economy) and smooth enough to guarantee regular learning.

Since HT-RGBF-PT-F1 is experimentally the best architecture and recipe on Trackmania, we finally tested it on Cyberpunk 2077, a first-person shooter video game (while Trackmania is a third-person racing game).
This model, indicated as CP-HT-RGBF-PT-F1* in the Table, yields $\mathrm{Acc.}=0.911$ and $F_1^{\mathrm{macro}}=0.617$ (compared to $0.938$ and $0.793$ for Trackmania, respectively).
In other words, results on Cyberpunk 2077 are slightly worse than those achieved on Trackmania.
This result appears reasonable given the higher complexity of Cyberpunk's visual and action scenarios (shooting and jumping integrated with base actions) and the sheer diversity in movement when comparing it to a constrained racing game like Trackmania.

\subsection{Qualitative analysis}
\label{sec:qualitative-analysis}
\begin{figure}[!tbp]
    \centering
    \begin{subfigure}[t]{0.49\textwidth}
    \includegraphics[width=\linewidth]{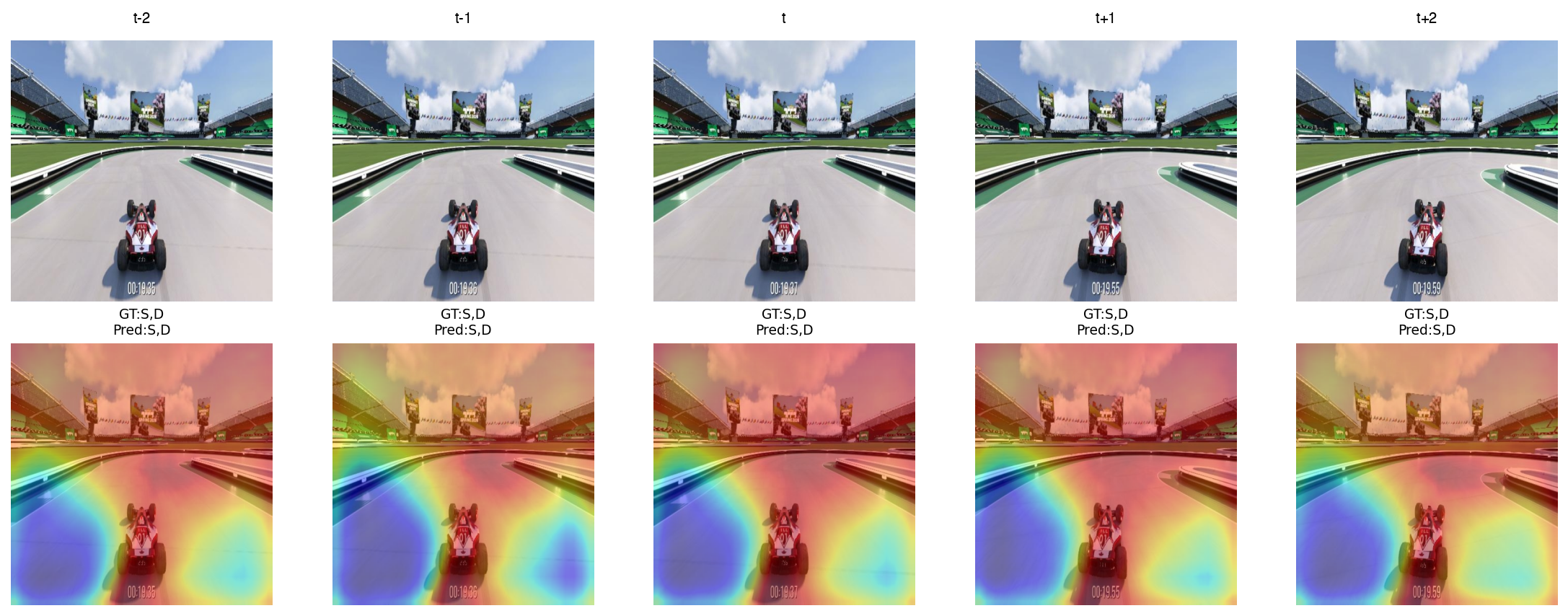}
    \caption{Braking (skid marks seen) as the vehicle approaches a visible curve ($S,D$).}
    \end{subfigure}
    \begin{subfigure}[t]{0.49\textwidth}
    \includegraphics[width=\linewidth]{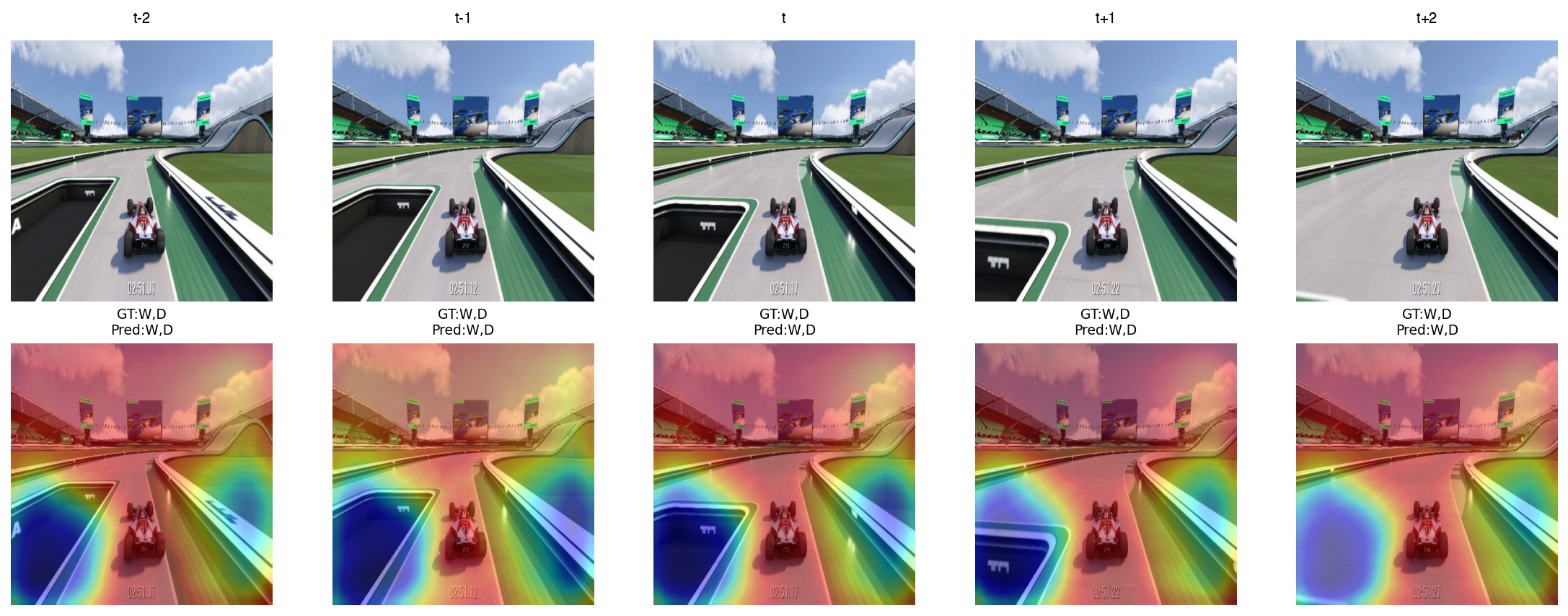}
    \caption{Forward-right on observing the slight wheel angle change and inertia ($W,D$).}
    \end{subfigure}
    \caption{Correctly predicted Trackmania sequences for HT-RGBF-PT-F1 (attention in the lower row). Pred matches GT for the supervised frame $t$. Better seen at 6x zoom.}
    \label{fig:qualitative-trackmania-correct}
\end{figure}

\begin{figure}[!tbp]
    \centering
    \begin{subfigure}[t]{0.49\textwidth}
    \includegraphics[width=\linewidth]{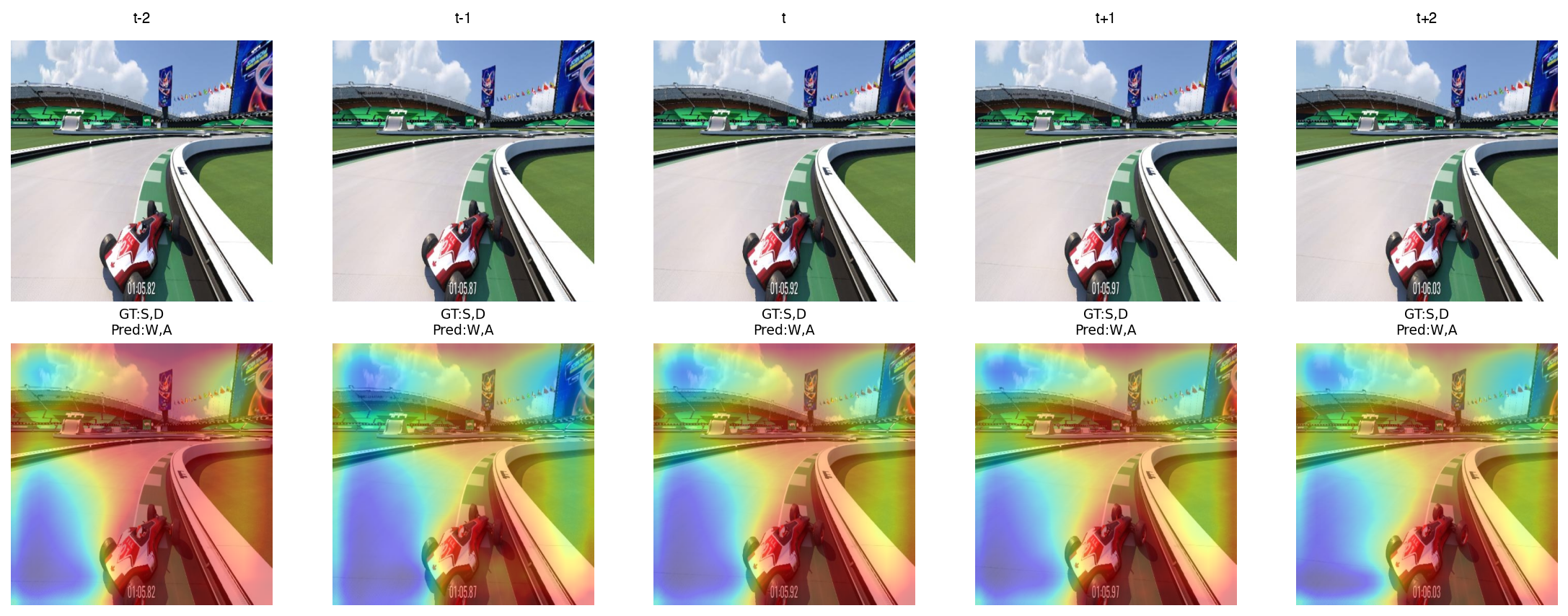}
    \caption{Reverse-right $(S,D)$ is confused with forward-left $(W,A)$ while fixing barrier collision.}
    \end{subfigure}
    \begin{subfigure}[t]{0.49\textwidth}
    \includegraphics[width=\linewidth]{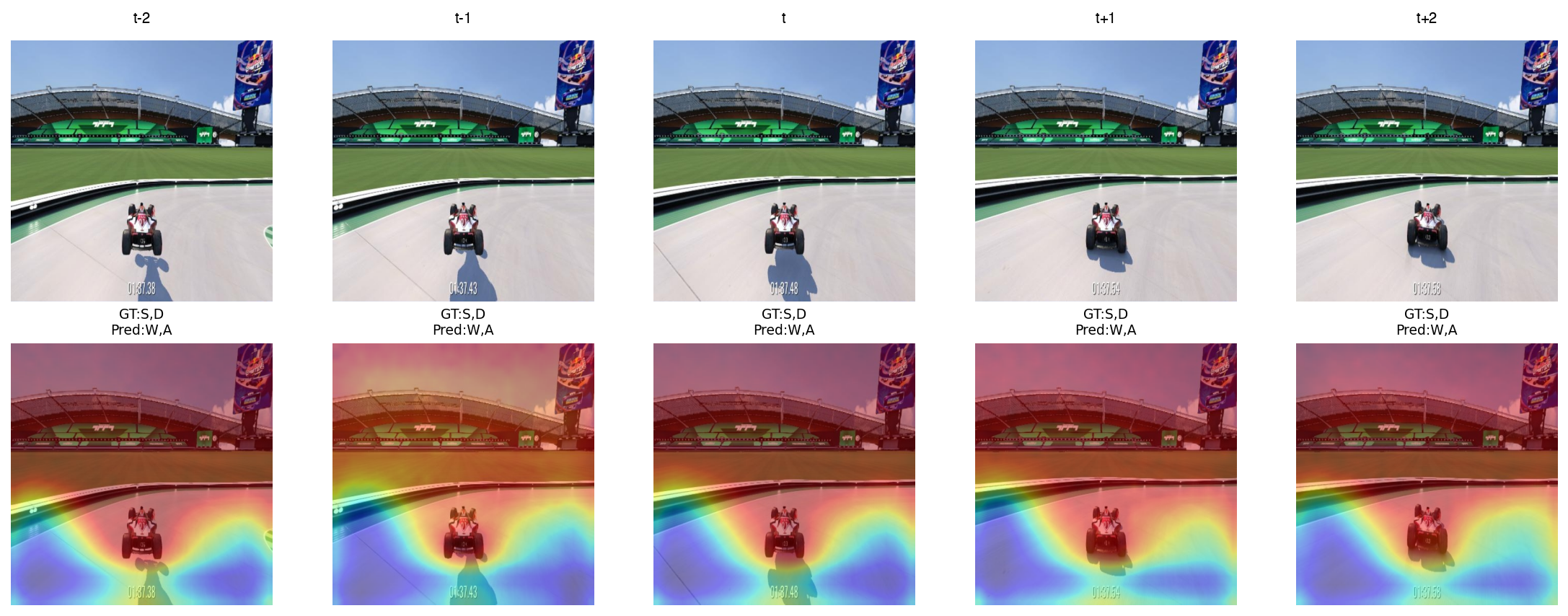}
    \caption{Airborne braking is omitted as it is ambiguous from the visual evidence.}
    \end{subfigure}
    \caption{Incorrectly predicted Trackmania sequences for HT-RGBF-PT-F1 (attention in the lower row). Reverse-right ($S,D$) is predicted as forward-left ($W,A$), indicating lack of 3D understanding, while airborne braking and steering ($S,D$) is predicted as ($W,A$) as it is visually ambiguous. Better seen at 6x zoom.}
    \label{fig:qualitative-trackmania-errors}
\end{figure}

\begin{figure}[!tbp]
    \centering
    \begin{subfigure}[t]{0.49\textwidth}
    \includegraphics[width=\linewidth]{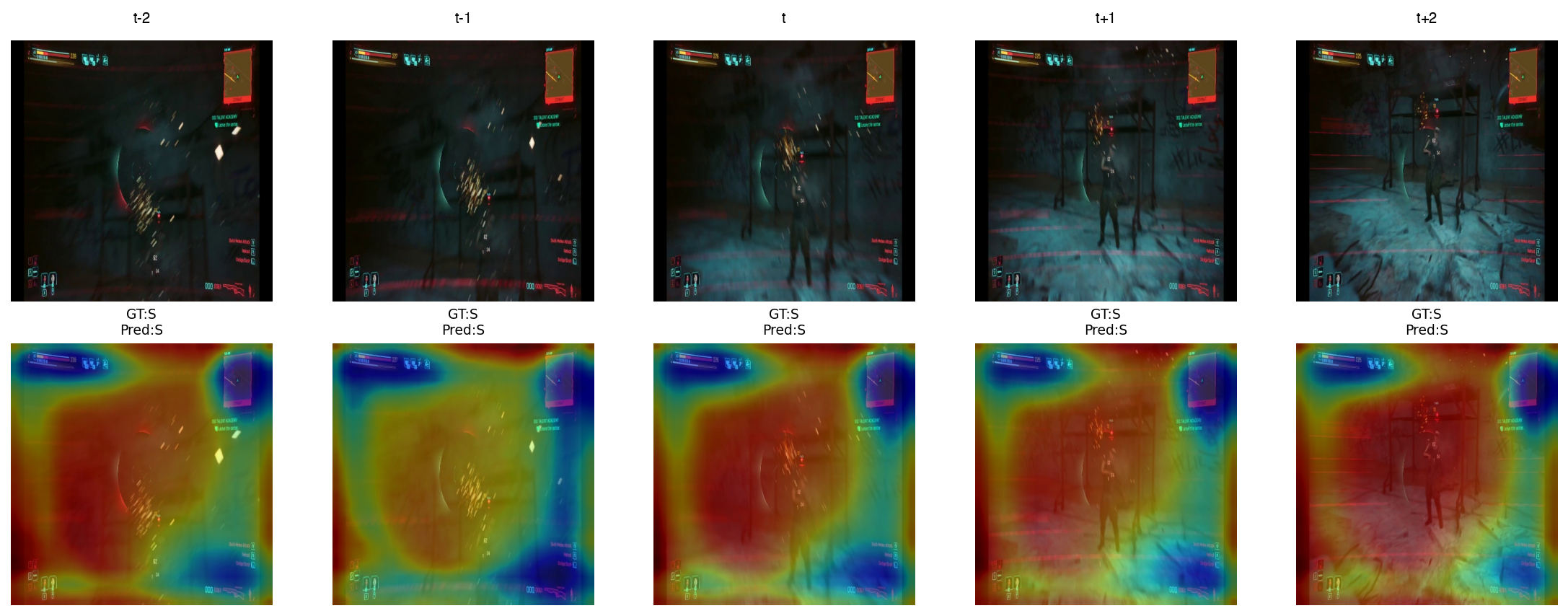}
    \caption{Backward movement recovered during combat ($S$).}
    \end{subfigure}
    \begin{subfigure}[t]{0.49\textwidth}
    \includegraphics[width=\linewidth]{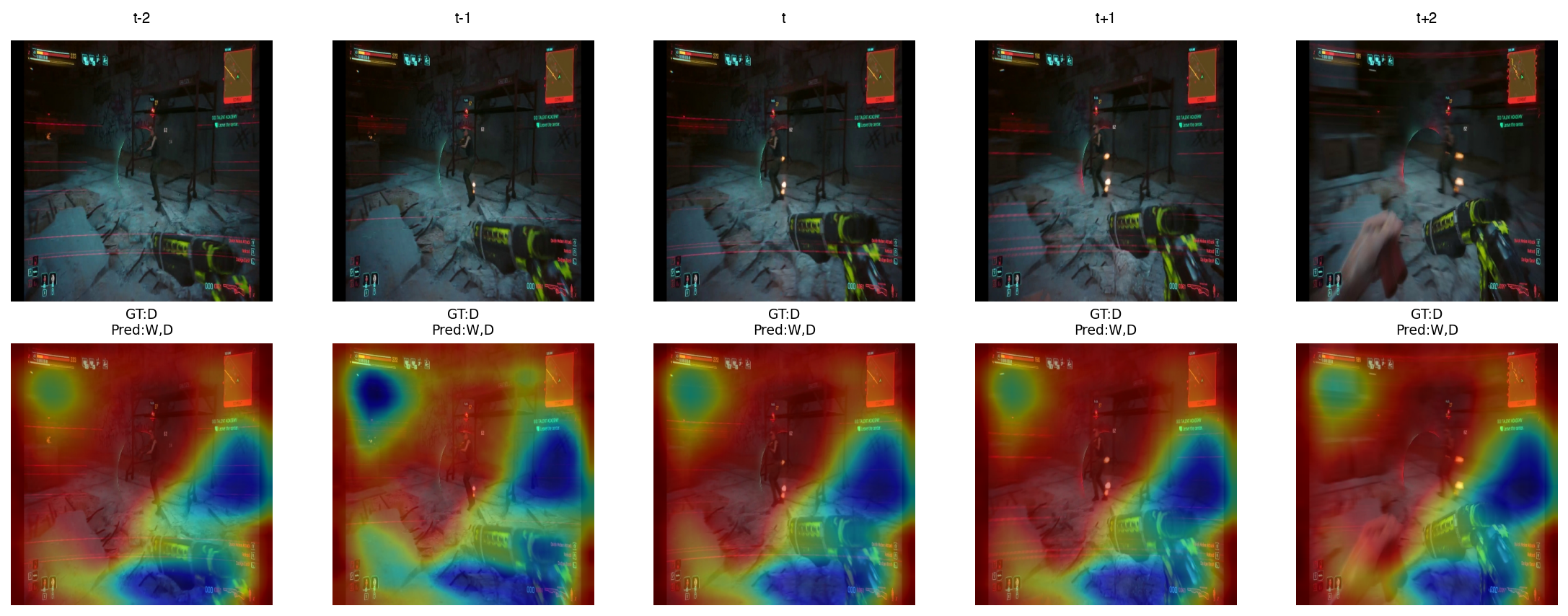}
    \caption{Right movement ($D$) recovered with a false forward prediction ($W,D$).}
    \end{subfigure}
    \caption{Correctly classified Cyberpunk 2077 sequences for CP-HT-RGBF-PT-F1 (attention in the lower row). Exact backward-motion match and partial right-movement match after being shot. Better seen at 6x zoom.}
    \label{fig:qualitative-cyberpunk-correct}
\end{figure}

\begin{figure}[!tbp]
    \centering
    \begin{subfigure}[t]{0.49\textwidth}
    \includegraphics[width=\linewidth]{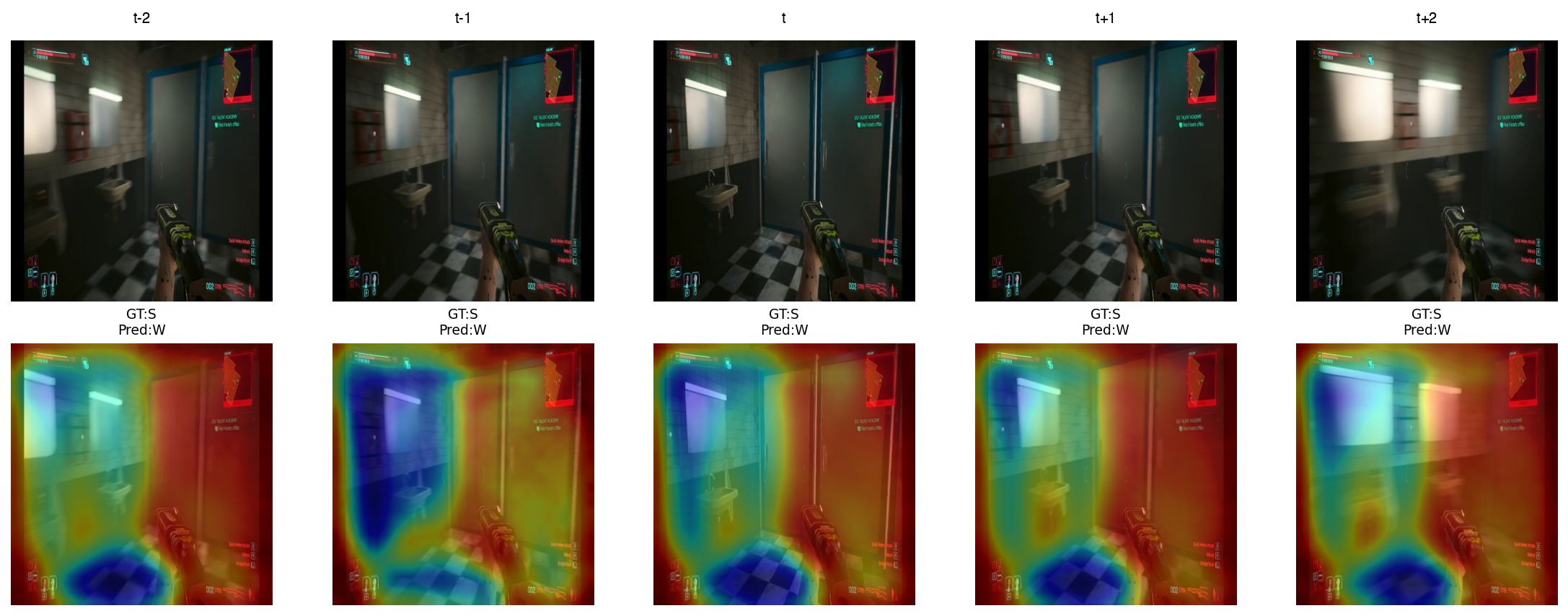}
    \caption{Backward movement ($S$) confused with forward movement ($W$).}
    \end{subfigure}
    \begin{subfigure}[t]{0.49\textwidth}
    \includegraphics[width=\linewidth]{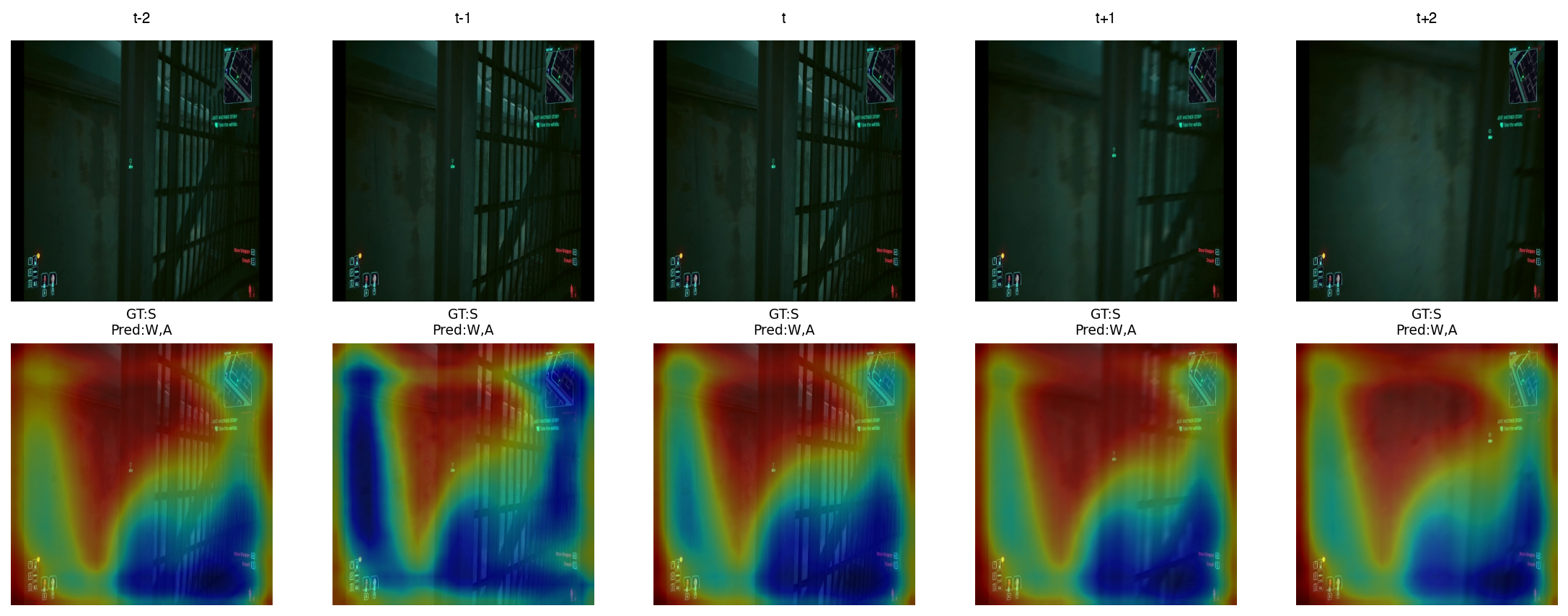}
    \caption{Backward movement ($S$) confused with forward-left ($W,A$).}
    \end{subfigure}
    \caption{Incorrectly classified Cyberpunk 2077 sequences for CP-HT-RGBF-PT-F1 (attention in the lower row). Both examples invert backward motion; in the second a false left movement is also predicted. Better seen at 6x zoom.}
    \label{fig:qualitative-cyberpunk-errors}
\end{figure}

To better understand the reasons for failure or success of the trained IDMs, we inspect sequences that show small, medium or large key-press reconstruction errors.
For each game, we show here two pairs of typical correct/incorrect predictions from the validation dataset using HT-RGBF-PT-F1 on Trackmania (Figures~\ref{fig:qualitative-trackmania-correct} and~\ref{fig:qualitative-trackmania-errors}) and CP-HT-RGBF-PT-F1 on Cyberpunk 2077 (Figures~\ref{fig:qualitative-cyberpunk-correct} and~\ref{fig:qualitative-cyberpunk-errors}).
Each example shows the $T_{in}=5$ frame sequence from $t-2$ to $t+2$ (top) and spatially aligned attention overlays on it (bottom).
After RGB and optical flow input concatenation, cross-attention is averaged over 8 heads and min-max normalized; the resulting map is overlaid on each frame to visualize its spatial contribution to the center-frame prediction.
GT and Pred in each panel denote the active ground-truth and 0.5 thresholded predicted keys.

\subsubsection{Trackmania} For correct predictions (Fig.~\ref{fig:qualitative-trackmania-correct}), predicted keys are supported by visual evidence that persists across the five frames: the skid marks support braking and right steering, while wheel pose and track-relative motion support forward-right prediction.
Notice also that the attention map often focuses on features that change with the camera motion, supporting the importance of motion flow in input.
The failure cases (Fig.~\ref{fig:qualitative-trackmania-errors}) reflect two ambiguities: when the vehicle is stuck, $3^{\mathrm{rd}}$-person camera motion makes reverse-right appear as forward-left, yielding $W,A$; airborne braking has no immediate visual effect and is therefore missed.
These cases motivate explicit separation of ego- and camera motion and longer temporal context for actions with delayed visual effects.

\subsubsection{Cyberpunk} Figures~\ref{fig:qualitative-cyberpunk-correct} and~\ref{fig:qualitative-cyberpunk-errors} show the corresponding Cyberpunk 2077 examples. The success case recovers sustained backward motion despite transient combat effects.
In the partial match, lateral motion supports $D$, but weapon and hit-response motion may cause the false $W$ prediction. This perhaps is the general policy of forward movement being applied. 
Both failures invert $S$ to $W$: one despite newly revealed fixed room features (white lights), and the other in a visually sparse corridor while also adding $A$.
They indicate that even optical flow cannot fully disentangle player from camera motion (which must indeed be performed by the IDM) and combat-specific motion in $1^{st}$-person scenes.

\section{Discussion and Conclusion}

\subsection{Positive Results and Design Implications}
Our main finding is that the spatiotemporal ViT outperforms traditional CNNs, while the HT architecture coupled with optical flow and contrastive initialization forms the strongest combination studied.
Our intuition is that the optical flow helps the model disentangle complex motion features (such as those created by camera motion/inertia while the playable character moves) while contrastive initialization helps build a more reliable embedding space- especially when training from scratch.
The embeddings learned on a target game mechanic (like navigation) or environment are likely more focused than general pretrained embeddings, as shown in the Appendix when comparing against a model trained with frozen DINOv3. 
Our experiments confirm that using accuracy $\mathrm{Acc.}$ in evaluation can hide minority-key errors because inactive (or always-on) keys may dominate it.
Further experiments in the Appendix seem to indicate that a balanced metric should be used for training as well (in fact we use a soft-F1 instead of BCE loss), but the numerical evidence is not sufficient to fully support this statement: more tests are needed to confirm it.
Our findings also indicate (see Appendix) that higher frame resolution provides small (yet consistent) improvements in performance, at the cost of higher compute.
Nonetheless, we decide to (and suggest future researchers to) work at the highest resolution to standardize subsequent models and avoid a resolution bottleneck in detail-heavy games like Cyberpunk.


\subsection{Limitations and Future Directions}
When compared against models trained for BC (and thus learning a reliable, average policy to predict the \emph{next} key-presses) on the same game, we found that our IDMs achieve only slightly superior performance.
For instance, a CNN model successfully trained to play Trackmania typically achieves $F_{1,W}\sim0.9$ and $F_{1,A} \sim F_{1,D}\sim0.8$, whereas our best IDM (see Table~\ref{tab:results-ablations-main-copy}) achieves $0.920$, $0.869$, and $0.870$ on the same metrics.
Our interpretation is that the IDM may partly learn the dataset's average policy rather than a precise visual-to-key mapping.
This interpretation requires further validation but motivates larger, more diverse datasets, as in VPT and D2E, to avoid regression toward the average policy. However, scale alone cannot remove visual ambiguity: gameplay projects 3D motion into 2D, while the camera moves independently.
Optical flow measures displacement, not its cause; a reversing vehicle can coincide with forward camera motion, while weapons and HUD overlays are independent of navigation.
Explicit 3D scene and ego-motion modeling may therefore be required to disentangle these dynamics and further improve IDM performance. This may aid the learning of player motion (and thus the reconstruction of the controlling key-presses) independently from that of the camera.

We also highlight that certain key-presses produce little or no immediate visual evidence, making reconstruction ambiguous from pixels alone (Figures~\ref{fig:qualitative-trackmania-errors}
and~\ref{fig:qualitative-cyberpunk-errors}). For example, steering into a wall may leave vehicle position and orientation unchanged except for a subtle wheel-angle change. Such a press is irrelevant only if it has no later consequence: inertia, limited-resource consumption, or adding an item to an inventory can alter latent game state without becoming visible within a $T$-frame context. Recovering these partially observable actions, when possible, may require a longer context, explicit state memory, and high resolution to capture minute visual details.
These aspects should be taken into careful consideration for future research in this direction.

\clearpage
\bibliographystyle{splncs04}
\bibliography{main}
\clearpage

\appendix
\renewcommand{\theHsection}{appendix.\Alph{section}}
\section*{Appendix (Supplementary Material)}

\section{Ablation study (secondary factors)}
\label{sec:ablation_secondary}

As explained in Section~\ref{sec:ablation_main}, we detail here additional ablation studies on factors that experimentally had minor, no, or negative impact on the model performance. These are: (i) frame resolution, (ii) pretrained embeddings, and (iii) loss functions.

\subsubsection{Frame Resolution:}
We evaluated a CNN variant, named CNN-128, that takes $128 \times 128$ input frames instead of the $512 \times 512$ frames used by CNN-512.
The architectures are otherwise identical, and both models have $16{,}685$ parameters; the comparison therefore isolates input resolution.
Rows 1 and 2 of Table~\ref{tab:results-ablations-secondary} compare the performance of the two architectures, showing a small (yet consistent) advantage for the higher resolution input frames, as expected.
This suggests working at the highest possible resolution, without (on the other hand) increasing the frame resolution beyond a certain limit when compute is an important factor.

\subsubsection{Pretrained Embeddings:}
We evaluated a ViT variant, named ViT-DINO-512, taking in input $512\times512$ (like ViT-E2E-512) and computing embeddings on the $16\times16$ patches using DINOv3~\cite{simeoni2025dinov3}.
While ViT-E2E-512 learns its patch projection end-to-end, the DINOv3 ViT-S/16 backbone in ViT-DINO-512 is frozen, removing 1 class and 4 register tokens before the common encoder.
Rows 3 and 4 of Table~\ref{tab:results-ablations-secondary} compare the performance of the two models trained with the exact same recipe and show no clear advantage for DINOv3 embeddings over the ones that are learned end-to-end.
This result seems to discourage the adoption of embedders pretrained on real-world data for video game scenarios, where the visual features may be largely different from typical scenes observed in the real world.
Since numerical differences are small, however, such a conclusion should be taken with a grain of salt: more experiments on different seeds and different games may be needed to draw a reliable conclusion.

\subsubsection{Loss Functions:}
We evaluated the effect of adopting different loss functions in training on the HT architecture.
In the ablation study, we compare this against the traditional BCE loss and an adaptive asymmetric loss (denoted as BCE${^\text{A}}$), which penalizes false predictions ($FP_k$ and $FN_k$) more for rarer keys: the BCE loss is multiplied by a factor of $25$ when the key frequency is $<0.10$, $20$ for frequency in the $[0.10,0.50)$ interval, and $5$ otherwise.
The last row of Table~\ref{tab:results-ablations-secondary} shows the corresponding model (HT-RGB-ASYM).
Also in this case, the performance differences for the same model trained with the same recipe and different loss functions are minor and therefore more experiments on different seeds and different games may be needed to draw a reliable conclusion.
Nonetheless, the metrics suggest a slight advantage for the soft-F1 loss, especially for the rarer $Sp$ key, as introduced in the main paper. This explains why we adopted it as our preferred choice.

\begin{table}[!t]
\centering
\caption{Experiment configurations for Tables~\ref{tab:results-ablations-main-copy} and~\ref{tab:results-ablations-secondary}. E2E is end-to-end; $d/L/H$ is transformer embedding dimension, number of blocks, and number of attention heads.}
\label{tab:architectures}
{
\scriptsize
\begin{tabular}{@{}p{0.22\textwidth}p{0.51\textwidth}p{0.24\textwidth}@{}}
Experiment & Configuration & Purpose \\
\midrule
CNN-128 & $128^2$, 15-channel RGB, soft-F1 & resolution baseline \\
CNN-512 & $512^2$, otherwise CNN-128 & resolution ablation\\
\midrule
ViT-E2E-512 & $512^2$, RGB, $d/L/H=384/2/4$, soft-F1 & ViT baseline\\
ViT-DINO-512 & ViT-E2E with DINOv3 features & spatial-pretraining ablation\\
\midrule
HT-RGB-E2E & $512^2$, 15-channel RGB, $d/L/H=256/4/8$, soft-F1 & HT baseline\\
HT-RGB-BCE & HT-RGB-E2E with BCE & loss ablation \\
HT-RGB-ASYM & HT-RGB-E2E with BCE$^{\text{A}}$ & loss ablation\\
HT-RGBF-E2E & HT-RGB-E2E with 8 flow channels (23 total) & flow ablation\\
\midrule
HT-RGBF-PT-F1 & HT-RGBF with 20-epoch contrastive pretraining and 25-epoch soft-F1 supervision & final recipe \\
CP-HT-RGBF-PT-F1 & HT-RGBF-PT-F1 applied to Cyberpunk 2077 & second-game application \\
\end{tabular}
}
\end{table}


\begin{table}[tb]
\centering
\caption{Accuracy Acc. and $F_1$ metrics on the Trackmania test split (for the secondary ablations reported in Table~\ref{tab:architectures});
$20+25$ denotes 20 contrastive and 25 supervised epochs.
\textbf{Bold} and \underline{underlined} mark the highest and second-highest distinct values in each metric column, respectively.}
\label{tab:results-ablations-secondary}
\resizebox{\textwidth}{!}{%
\begin{tabular}{cccccccc|cccccccc}
\multicolumn{1}{c}{\multirow{2}{*}{Model}} & \multicolumn{7}{c|}{Configuration} & \multicolumn{1}{c}{\multirow{2}{*}{Acc.}} & \multicolumn{7}{c}{$F_1$} \\
& $T_{\mathrm{out}}$ & Arch. & Res. & Flow & Initialization & Loss & Epochs & & micro & macro & W & A & S & D & Sp \\
\midrule
CNN-128          & $5$ & CNN & $128^2$ & N & none & soft-F1 & 100 & .469 & .492 & .361 & .902 & .430 & .074 & .392 & .009 \\
CNN-512          & $5$ & CNN & $512^2$ & N & none & soft-F1 & 100 & .489 & .499 & .367 & .902 & .434 & .075 & .415 & .008 \\
\midrule
ViT-E2E-512      & $5$ & ViT & $512^2$ & N & none & soft-F1 & 100 & .887 & .817 & .637 & .907 & .698 & .567 & .755 & .256 \\
ViT-DINO-512     & $5$ & ViT & $512^2$ & N & DINOv3 & soft-F1 & 100 & .870 & .790 & .624 & .898 & .696 & .452 & .664 & .411 \\
\midrule
HT-RGB-E2E       & $1$ & HT  & $512^2$ & N & none & soft-F1 & 50 & \underline{.929} & \textbf{.880} & \textbf{.746} & \textbf{.921} & \textbf{.837} & \textbf{.648} & \underline{.823} & \textbf{.502} \\
HT-RGB-BCE       & $1$ & HT  & $512^2$ & N & none & BCE & 50 & \textbf{.930} & \underline{.875} & \underline{.730} & \underline{.919} & \underline{.828} & \underline{.642} & \textbf{.825} & .437 \\
HT-RGB-ASYM      & $1$ & HT  & $512^2$ & N & none & BCE$^{\mathrm{A}}$ & 50 & .922 & .861 & .719 & .916 & .801 & .625 & .796 & \underline{.456} \\
\end{tabular}
}
\end{table}

\section{Training and Validation Curves}
\label{app:training-validation-curves}

Figures~\ref{fig:appendix-supcon-curves} and~\ref{fig:appendix-supervised-curves} report training and validation losses and relevant metrics within the epoch budgets used in Table~\ref{tab:results-ablations}.
Reported checkpoints maximize validation macro-$F_1$ within these budgets.

\begin{center}
    \includegraphics[width=\textwidth]{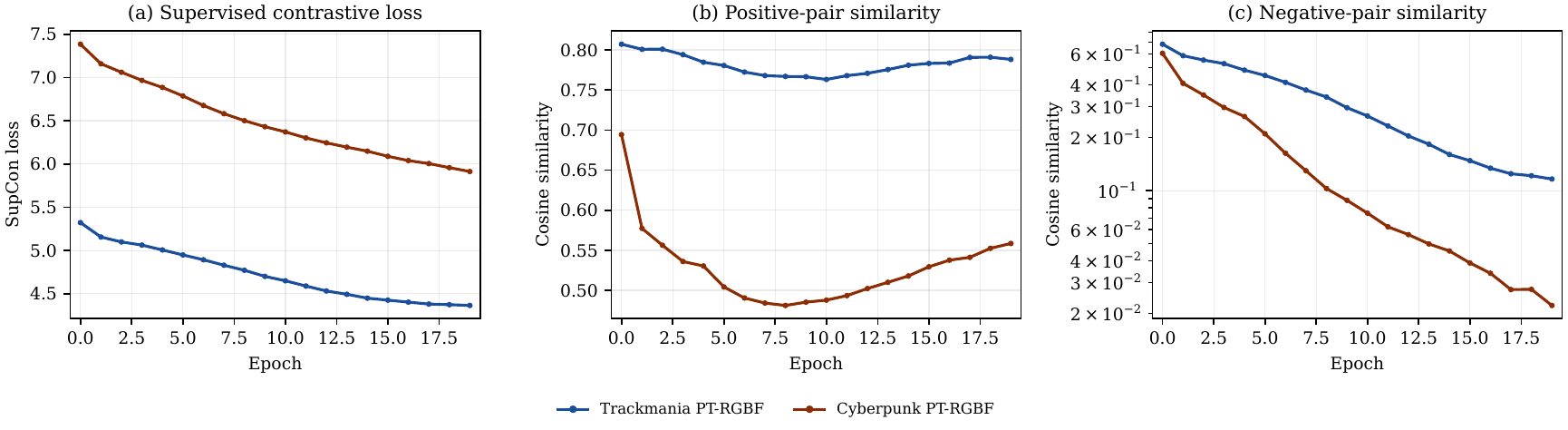}
    \captionof{figure}{Contrastive-pretraining curves for the RGB+flow HT encoders on Trackmania and Cyberpunk 2077: (a) supervised contrastive loss, (b) positive-pair cosine similarity, and (c) negative-pair cosine similarity over the 20-epoch pretraining budget.}
    \label{fig:appendix-supcon-curves}
\end{center}

\begin{center}
    \includegraphics[width=\textwidth]{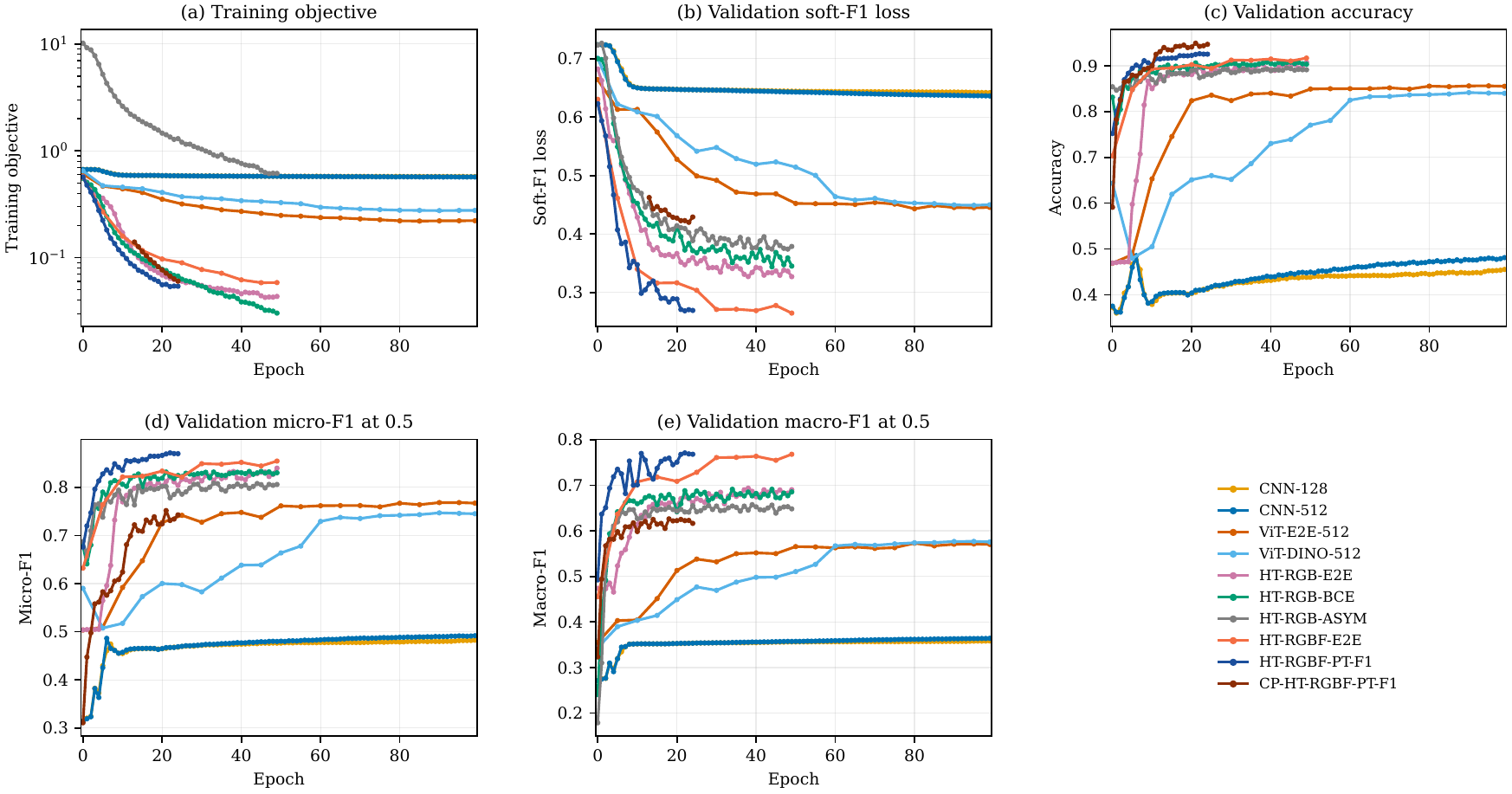}
    \captionof{figure}{Supervised-training curves within the epoch budgets reported in Table~\ref{tab:results-ablations}: (a) training objective (loss), (b) validation soft-$F_1$ loss, (c) validation accuracy, (d) validation micro-$F_1$, and (e) validation macro-$F_1$. Thresholded metrics use 0.5. Pretrained models begin supervised epoch 0 after 20 contrastive epochs; training-objective values are comparable only between runs using the same loss (asymmetric loss scales differently as noted in Section~\ref{training_details}).}
    \label{fig:appendix-supervised-curves}
\end{center}
\clearpage

\end{document}